\def\year{2018}\relax
\documentclass[letterpaper]{article} 
\usepackage{aaai18}  
\usepackage{times}  
\usepackage{helvet}  
\usepackage{courier}  
\usepackage{url}  
\usepackage{graphicx}  
\usepackage{stfloats}

\usepackage{amssymb}
\usepackage{amsmath}
\usepackage{mathabx}
\usepackage{tikz}
\usetikzlibrary{calc}
\usepackage{algorithm}
\usepackage[noend]{algpseudocode}
\usepackage{xr}

\newtheorem{theorem}{Theorem}

\newtheorem{lemma}[theorem]{Lemma}
\newtheorem{definition}{Definition}
\newtheorem{example}{Example}
\newtheorem{prop}{Proposition}

\newtheorem{property}{Property}

\DeclareMathOperator*{\argmin}{arg\,min}
\DeclareMathOperator*{\argmax}{arg\,max}

\usepackage{placeins}

\newcommand{\T}{\mathcal{T}}
\newcommand{\A}{\mathcal{A}}

\begin{document}
%
\title{Computational Results for Extensive-Form Adversarial Team Games}
\author{Andrea Celli \textnormal{and} Nicola Gatti\\ ~ Politecnico di Milano\\ ~ Piazza Leonardo da Vinci, 32\\ ~ Milano, Italy\\ ~ \{andrea.celli, nicola.gatti\}@polimi.it}

\maketitle

\begin{abstract}
We provide, to the best of our knowledge, the first computational study of extensive-form adversarial team games. These games are sequential, zero-sum games in which a team of players, sharing the same utility function, faces an adversary. We define three different scenarios according to the communication capabilities of the team. In the first, the teammates can communicate and correlate their actions both \emph{before} and \emph{during} the play. In the second, they can only communicate \emph{before} the play. In the third, \emph{no} communication is possible at all. We define the most suitable solution concepts, and we study the inefficiency caused by partial or null communication, showing that the inefficiency can be arbitrarily large in the size of the game tree. Furthermore, we study the computational complexity of the equilibrium-finding problem in the three scenarios mentioned above, and we provide, for each of the three scenarios, an exact algorithm. Finally, we empirically evaluate the scalability of the algorithms in random games and the inefficiency caused by partial or null communication.
\end{abstract}

\section{Introduction}
\vspace{-0.1cm}
The design of algorithms for strategic settings has been a central problem in Artificial Intelligence for several years, with the aim of developing agents capable of behaving optimally when facing strategic opponents. Many efforts have been made for 2-player games, e.g., finding a \emph{Nash} equilibrium~\cite{lemke64equilibrium,gatti2012} and, more recently, finding a \emph{Stackelberg} equilibrium~\cite{conitzer2006computing}. The study of this latter problem paved the way to the field of Security Games, which is, nowadays, one of the application fields of non-cooperative game theory with the highest social impact~\cite{tambe2011}.

Fewer results are known, instead, about games with more than 2 players---except for games with particular structures, e.g., congestion and polymatrix games~\cite{nisan2007algorithmic}. An interesting class of games widely unexplored in the literature is that one of adversarial team games. Here, a team of players with the same utilities faces an adversary~\cite{vonStengelKoller1997}. These games can model many realistic security scenarios and can provide tools to coordinate teammates acting strategically. Basilico \emph{et al.}~\shortcite{basilico2016} study the inefficiency a team can incur in normal-form games when teammates cannot correlate their strategies w.r.t. when they can. They also provide algorithms to approximate the \emph{Team-maxmin} equilibrium---introduced by von~Stengel and Koller~\shortcite{vonStengelKoller1997}---that is the optimal solution when correlation is not possible. Furthermore, it is known that finding a Team-maxmin equilibrium is \textsf{FNP}-hard and inapproximable in additive sense~\cite{hansen2008,borgs2010}.

When extensive-form games enter the picture, adversarial team games become much more intriguing, various forms of correlation being possible. Nevertheless, to the best of our knowledge, this game class is still unexplored in the literature. In the present paper, we focus on three main forms of correlation~\cite{forges1986}. In the first, \emph{preplay} and \emph{intraplay} communication is possible, corresponding to the case in which a \emph{communication device} receives \emph{inputs} from the teammates about the information they observe during the play, and sends them \emph{recommendations} about the action to play at each information set. In the second, \emph{only preplay} communication among the teammates is possible, corresponding to the case in which a \emph{correlation device} communicates a plan of actions to each teammate before playing the game.\footnote{With only preplay correlation, three solution concepts are known: Normal-Form, Extensive-Form, and Agent-Form Correlated Equilibrium. The spaces of players' strategies achievable with the three correlation devices are the same, while the players' incentive constraints are different (even if it is not known whether the spaces of the outcomes for the three equilibria in  adversarial team games are different). The complexity of computing the equilibrium maximizing the team's utility in adversarial team games with at least 2 teammates is \textsf{NP}-hard for all the three equilibria~\cite{vonStengel2008}. In our paper, we focus on the first one.} Finally, in the third, \emph{no communication} is possible\footnote{This setting is frequent in practice. Consider, for example, a security problem where a set of defensive resources from different security agencies are allocated to protect an environment at risk but, due to organizational constraints, they are not able to share information with each other. The resources have the same goal (i.e., to secure the environment) but cannot coordinate strategy execution. The same happens when a set of resources has to operate in stealth mode.}. 

\textbf{Original contributions}. We formally define game models capturing the three aforementioned cases and the most suitable solution concepts  (trivially, the Team-maxmin equilibrium in the third setting). Furthermore, we define three inefficiency indices for the equilibria, capturing: the inefficiency caused by using a correlation device in place of a communication device, the inefficiency caused by not using any form of communication in place of a communication device, and the inefficiency caused by not using any form of communication in place of a correlation device. We provide lower bounds to the worst-case values of the inefficiency indices, showing that they can be arbitrarily large. 

Furthermore, we study the computational complexity of the problems of finding the three equilibria with the different forms of correlation, and we design, for each equilibrium, an exact algorithm. We show that when a communication device is available, an equilibrium can be computed in polynomial time (even in the number of players) by a 2-stage algorithm. In the first stage, the game is cast into an auxiliary 2-player equivalent game, while, in the second stage, a solution is found by linear programming. When a correlation device is available, the problem can be easily shown to be \textsf{FNP}-hard. In this case, we prove that there is always an equilibrium with a small (linear) support, and we design an equilibrium-finding algorithm, based on a classical column-generation approach, that does not need to enumerate an exponential space of actions before its execution. Our algorithm exploits an original hybrid representation of the game combining both normal and sequence forms.
The column-generation oracle is shown to deal with an \textsf{APX}-hard problem, with an upper approximation bound decreasing exponentially in the depth of the tree. We also provide an approximation algorithm for the  oracle that matches certain approximation guarantees on a subset of instances. When no communication is possible, the equilibrium-finding problem can be easily shown to be \textsf{FNP}-hard. In this case, the problem can be formulated as a non-linear programming problem and solved by resorting to global optimization tools. 

Finally, we empirically evaluate the scalability of our algorithms in random game instances. We also evaluate the inefficiency for the team of not adopting a communication device, showing that, differently from the theoretical worst-case bounds, the empirical inefficiency is extremely small.
\section{Preliminaries}
A \emph{perfect-information extensive-form game}~\cite{shoham2009multiagent} is a tuple $(N,A,V,L,\iota,\rho,\chi,U)$, where: $N$ is a set of $n$ players, $A$ is a set of actions, $V$ is the set of nonterminal decision nodes, $L$ is the set of terminal (leaf) nodes, $\iota: V\rightarrow N$ is a function returning the player acting at a given decision node, $\rho:V\rightarrow 2^{A}$ is the action function---assigning to each choice node a set of available actions---, $\chi: V\times A\rightarrow V\cup L$ is the successor function, and $U=\{U_1,U_2,\ldots,U_n\}$ is the set of utility functions in which $U_i:L\rightarrow \mathbb{R}$ specifies utilities over terminal nodes for player $i$. Let $V_i$ be the inclusion-wise maximal set of decision nodes such that, for all $x\in V_i$, $i=\iota(x)$. Then, an \emph{imperfect-information extensive-form game} is a tuple $(N,A,V,L,\iota,\rho,\chi,U,H)$, where $(N,A,V,L,\iota,\rho,\chi,U)$ is an extensive-form game with perfect information and $H=\{H_1,H_2,\ldots,H_n\}$ is the set of information sets, in which $H_i$ is a partition of $V_i$ such that, for any $x_1,x_2\in V_i$, $\rho(x_1)=\rho(x_2)$ whenever there exists a $h\in H_i$ where $x_1\in h$ and $x_2\in h$. As usual in game theory, we assume, for each $a\in A$, there is only one $h$ s.t. $a \in \rho (h)$. We focus on games with \emph{perfect recall} where, for each player $i$ and each $h\in H_i$, decision nodes belonging to $h$ share the same sequence of moves of player $i$ on their paths from the root.

The study of extensive-form games is commonly conducted under other equivalent representations. The \emph{normal form} is a tabular representation in which player~$i$'s actions are plans $p \in P_i$, specifying a move at each information set in~$H_i$,  and player~$i$'s utility is $U_i':P_1\times\ldots\times P_n\rightarrow \mathbb{R}$ s.t. $U_i'(p_1,\ldots,p_n)=U_i(l)$, where $l\in L$ is the terminal node reached when playing plan profile $(p_1,\ldots,p_n)$. Basically, in the normal-form representation, players decide their behavior in the whole game \emph{ex ante} the play. The \emph{reduced} normal form is obtained by deleting replicated strategies from the normal form. However, the size of the reduced normal form is exponential in the number of information sets. A \emph{mixed strategy} $\sigma_i$ of player $i\in N$ is a probability distribution on her set of pure strategies $P_i$. In the \emph{agent form}---whose definition is omitted due to reasons of space, see~\cite{selten1975}---, players play \emph{behavioral strategies}, denoted by $\pi_i(h,a)$, each specifying a probability distribution over the actions~$\rho(h)$ available at information set~$h$ of player~$i$. Two strategies, even of different representations, are \emph{realization equivalent} if, for any strategy profile of the opponents, they induce the same probability distribution over the outcomes. In a finite perfect-recall game, any mixed strategy can be replaced by an equivalent behavioral one~\cite{kuhn1953}. 

Both normal and agent forms suffer from computational issues that can be overcome by using the \emph{sequence form}~\cite{vonStengel1996}, whose size is linear in the size of the game tree. A sequence for player $i$, defined by a node $x$ of the game tree, is the subset of $A$ specifying player $i$'s actions on the path from the root to $x$. We denote the set of sequences of player~$i$ by~$Q_i$, these are the sequence-form actions of player~$i$. A sequence is said \textit{terminal} if, together with some sequences of the other players, leads to a terminal node. 
Moreover, we denote by~$q_\emptyset$ the fictitious sequence leading to the root node and with $qa\in Q_i$ the \textit{extended} sequence obtained by appending $a\in A$ to sequence $q\in Q_i$. The sequence-form strategy, said \emph{realization plan}, is a function $r_i:Q_i\rightarrow \mathbb{R}$ associating each sequence $q\in Q_i$ with its probability of being played. A well-defined sequence-form strategy is such that, for each $i\in N$, $r_i(q_\emptyset)=1$, for each $h$ and sequence $q$ leading to $h$, $-r_{i}(q)+\sum_{a\in\rho(h)} r_{i}(qa)=0$ and $r_i(q)\geq 0$. Constraints are linear in the number of sequences and can be written as $F_i \, r_i=f_i$, where $F_i$ is an opportune matrix and $f_i$ is an opportune vector. 
The utility function of player $i$ is represented as an $n$-dimensional matrix defined only for profiles of terminal sequences leading to a leaf. With a slight abuse of notation, we denote it by $U_i$.

A Nash equilibrium (NE), whose definition does not depend on the representation of the game, is a strategy profile in which no player can improve her utility by deviating from her strategy once fixed the strategies of all the other players.
\section{Extensive-Form Adversarial Team Games, Equilibria, and Inefficiency}

We initially provide the formal definition of a team.

\begin{definition}[Team]\label{def:team}
Given an extensive-form game with imperfect information $(N, A, V, L, \iota, \rho, \chi, U, H)$, a team $\mathcal{T}$ is an inclusion-wise maximal subset of players $T\subseteq N$ such that, for any $i,j\in T$, for all $l\in L$, $U_{i}(l)=U_{j}(l)$.
\end{definition}

We denote by $H_\mathcal{T}$ the set $\bigcup_{i\in T} H_{i}$ and by $A_\mathcal{T}$ the set of actions available at the information sets in $H_\T$. An extensive-form team game (EF-TG) is  a generic extensive-form game where at least one team is present. Von~Stengel and Koller~\shortcite{vonStengelKoller1997} analyze zero-sum normal-form games where a single team plays against an adversary. We extend this game model to the scenario of extensive-form games. 

\begin{definition} [STSA-EF-TG] \label{def:sast_ef_tg}
  A zero-sum single-team single-adversary extensive-form team game (STSA-EF-TG) is a game $(N, A, V, L, \iota, \rho, \chi, U, H)$ in which:
  \begin{itemize}
    \item $N=T\cup\{n\}$, where set $T$ defines a team $\mathcal{T}$ (as in Definition~\ref{def:team}) and player $n$ is the adversary ($\mathcal{A}$);
    \item for each $l\in L$ it holds: $U_{\mathcal{A}}(l)=-(n-1)U_{\mathcal{T}}(l)$, where $U_\mathcal{T}$ denotes the utility of teammates and $U_\mathcal{A}$ that one of the adversary.
  \end{itemize}
\end{definition}
When the teammates have no chance to correlate their strategies, the most appropriate solution concept is the Team-maxmin equilibrium (TME). Formally, the TME is defined as $\arg \max_{r_1,\ldots, r_{n-1}} \min_{r_n} U_{\mathcal{T}} \prod_{i=1}^n r_i$. By using the same arguments used by von~Stengel and Koller~\shortcite{vonStengelKoller1997} for the case of normal-form games, it follows that also in extensive-form games a TME is unique except for degeneracy and it is the NE maximizing team's expected utility. Nevertheless, in many scenarios, teammates may exploit higher correlation capabilities. While in normal-form  games these capabilities reduce to employing a correlation device as proposed by~\cite{aumann1974}, in extensive-form games we can distinguish different forms of correlation. More precisely, the strongest correlation is achieved when teammates can communicate both before and during the execution of the game (preplay and intraplay communication), exchanging their private information by exploiting a \emph{mediator} that recommends actions to them. This setting can be modeled by resorting to a \emph{communication device} defined in a similar way to~\cite{forges1986}. A weaker correlation is achieved when teammates can communicate only before the play (preplay communication). This setting can be modeled by resorting to a \emph{correlation device} analogous to that one for normal-form games. We formally define these two devices as follows (as customary, $\Delta(\cdot)$ denotes the simplex over $\cdot$). 

\begin{definition}[Communication device]\label{def:communication_dev}
	A communication device is a triple $(H_\T,A_\T,R^{\textnormal{Com}})$ where $H_\T$ is the set of inputs (i.e., information sets) that teammates can communicate to the mediator, $A_\T$ is the set of outputs (i.e., actions) that the mediator can recommend to the teammates, and  $R^{\textnormal{Com}}:2^{H_\T}\times2^{A_\T}\rightarrow\Delta(A_\T)$ is the recommendation function that associates each information set $h\in H_\T$ with a probability distribution over $\rho(h)$, as a function of information sets previously reported by teammates and of the actions recommended by the mediator in the past.
\end{definition}
\begin{definition}[Correlation device]\label{def:correlation_device}
	A correlation device is a pair $(\{P_i\}_{i\in T},R^{\textnormal{Cor}})$. $R^{\textnormal{Cor}}:\bigtimes_{i\in T}P_i\rightarrow\Delta(\bigtimes_{i\in T}P_i)$ is the recommendation function which returns a probability distribution over the reduced joint plans of the teammates.
\end{definition}

Notice that, while a communication device provides its recommendations drawing actions from probability distributions during the game, a correlation device does that only before the beginning of the game. Resorting to these definitions, we introduce the following solution concepts.
\begin{definition}[Team-maxmin equilibrium variations]\label{def:corr_equilibria}
	Given a communication device---or a correlation device---for the team, a Team-maxmin equilibrium with communication device (TMECom)---or a Team-maxmin equilibrium with correlation device (TMECor)---is a Nash equilibrium in which all teammates follow their recommendations and, only for TMECom, report truthfully their information.
\end{definition}

Notice that in our setting (i.e., zero-sum games), both TMECom and TMECor maximize team's utility.
We state the following, whose proof is straightforward. 
\begin{property}[Strategy space]
The space of lotteries over the outcomes achievable by using a  communication device includes that one of the lotteries achievable by using a correlation device, that, in its turn, includes the space of the lotteries achievable without any device. 
\end{property}

Let $v_{\textnormal{No}}$, $v_{\textnormal{Com}}$, $v_{\textnormal{Cor}}$ be the utility of the team at, respectively, the TME, the TMECom and the TMECor. From the property above, we can easily derive the following. 
\begin{property}[Equilibria utility]
The game values obtained with the different solution concepts introduced above are such that $v_{\textnormal{Com}}\geq v_{\textnormal{Cor}} \geq v_{\textnormal{No}}$.
\end{property}

In order to evaluate the inefficiency due to the impossibility of adopting a communication or correlation device, we resort to the concept of Price of Uncorrelation ($PoU$), previously introduced in~\cite{basilico2016} as a measure of the inefficiency of the TME w.r.t. the TMECor in normal-form games. In these games, the $PoU$ is defined as the ratio between the utility given by the TMECor and the utility given by the TME, once all the team's payoffs are normalized in $[0,1]$. For extensive-form games, we propose the following variations of the $PoU$ to capture all the possible combinations of different forms of correlation.
\begin{definition}[Inefficiency indices]\label{def:pous}
$PoU_{\textnormal{Com/No}}=\frac{v_{\textnormal{Com}}}{v_{\textnormal{No}}}$, $PoU_{\textnormal{Cor/No}}=\frac{v_{\textnormal{Cor}}}{v_{\textnormal{No}}}$, $PoU_{\textnormal{Com/Cor}}=\frac{v_{\textnormal{Com}}}{v_{\textnormal{Cor}}}$.
\end{definition}

In perfect-information games all these indices assume a value of 1, the solution being unique unless degeneracy by backward induction. With imperfect information the indices can be larger than 1. In normal-form games, the tight upper bound to $PoU$ is $m^{n-2}$, where $m$ is the number of actions of each player and $n$ is the number of players~\cite{basilico2016}. Using a definition based on $m$ is not suitable for extensive-form games, where each player may have a different number of actions per node. Thus, we state the bounds in terms of $|L|$ (i.e., the number of terminal nodes). The following three examples provide lower bounds to the worst-case values of the indices, showing that the inefficiency may be arbitrarily large in $L$.
%
Initially, to ease the presentation, we define a specific type of team player that we call \textit{spy}.

\begin{definition}[Spy player]\label{def:spy}
	Player $i\in T$ is said to be a spy if, for each $h\in H_i$, $|\rho(h)|=1$ and $h$ is a singleton.
\end{definition}
A spy just observes the actual state of the game and her contribution to the play is only due to her communication capabilities. Notice that the introduction of a spy after decision nodes of the adversary does not affect the team's utility in a TMECor (the team's joint plans  are the same) but improves the team's capabilities, and final utility, in a TMECom. 

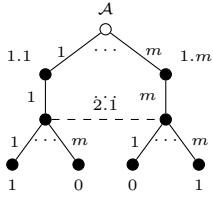
\begin{figure}
	\centering
	\begin{tikzpicture}[scale=0.4,font=\tiny]
			\tikzset{
				solid node/.style={circle,draw, inner sep=1.5, fill=black},
				hollow node/.style={circle,draw, inner sep=1.5}
				}
			\tikzstyle{level 1}=[level distance=15mm,sibling distance=20mm]
			\tikzstyle{level 2}=[level distance=15mm,sibling distance=23mm]
			\tikzstyle{level 3}=[level distance=15mm,sibling distance=11mm]
			\node(0)[hollow node, label=above:{$\mathcal{A}$}]{}
				child{node[solid node, label=above left:{$1.1$}]{} 
					child{node(a)[solid node, label=left:{$$}]{} 
						child{node(c)[solid node, label=below:{$1$}]{} edge from parent node[left]{$1$}}
						child {node[draw=none] {} edge from parent[draw=none] node{$\ldots$}}
						child{node(d)[solid node,label=below:{$0$}]{} edge from parent node[right]{$m$}}
						edge from parent node[left]{$1$}
					}
					edge from parent node[left]{$1$}
				}
				child{node[draw=none] {} 
					child {node[draw=none] {} edge from parent[draw=none] node{$\ldots$}}
					edge from parent[draw=none] node{$\ldots$}
				}
				child{node[solid node, label=above right:{$1.m$}]{} 
					child{node(b)[solid node]{} 
						child{node(e)[solid node,label=below:{$0$}]{} edge from parent node[left]{$1$}}
						child {node[draw=none] {} edge from parent[draw=none] node{$\ldots$}}
						child{node(f)[solid node,label=below:{$1$}]{} edge from parent node[right]{$m$}}
						edge from parent node[left]{$m$}
					}
					edge from parent node[right]{$m$}
				}
				;
				\draw[dashed](a)--node[above]{$2.1$}(b);
	\end{tikzpicture}
	\vspace{-0.4cm}
	\caption{A game with a spy used in Example~\ref{ex:comm_m}.}
	\label{fig:tree_comm_m}
	\vspace{-0.2cm}
\end{figure}

\begin{example}[Lower bound for worst-case $PoU_{\textnormal{Com/No}}$]\label{ex:comm_m}
Consider a STSA-EF-TG with $n$ players and $m\geq 2$ actions for each player at every decision node except for the first team player, who is a spy. The game tree is structured as follows (see Figure~\ref{fig:tree_comm_m} for the case with $n=3$).
\begin{itemize}
	\item The adversary plays first;
	\item then the spy observes her move;
	\item each one of the other teammates is assigned one of the following levels of the game tree and all her decision nodes are part of the same information set;
	\item $U_\T=1$ iff, for each $i\in T\backslash\{1\}$ and for each $h\in H_i$, the action chosen at $h$ is equal to the one selected by $\mathcal{A}$.
\end{itemize}
We have $v_{\textnormal{Com}}=1$, $v_{\textnormal{No}}=m^{2-n}$ and thus $PoU_{\textnormal{Com/No}}=m^{n-2}$. Since the tree structure is such that $|L|=m^{n-1}$ we obtain $PoU_{\textnormal{Com/No}}=|L|^{(1-\frac{1}{n-1})}$. Once $|L|$ is fixed, the inefficiency is monotonically increasing in $n$, but $n$ is upper bounded by $n=\log_2(|L|)+1$ (corresponding to the case in which each team player except the spy has the minimum number of actions, i.e., 2). It follows that, in the worst case w.r.t.~$n$, $PoU_{\textnormal{Com/No}}=\frac{|L|}{2}$.
\end{example}

\begin{example}[Lower bound for worst-case  $PoU_{\textnormal{Cor/No}}$] \label{ex:cor_m}
Consider a STSA-EF-TG with $n$ players and $m$ actions at each of their decision nodes, in which each level of the game tree is associated with one player and forms a unique information set. $U_\T=1$ iff all the teammates choose the same action of the adversary, who plays first. This case corresponds to the worst case for $PoU$ in normal-form games. Here we formulate the bound in terms of $|L|$. We have $v_{\textnormal{No}}=m^{1-n}$ and $v_{\textnormal{Cor}}=1/m$. It follows that $PoU_{\textnormal{Cor/No}}=m^{n-2}$. This time, $|L|=m^n$ and thus $PoU_{\textnormal{Cor/No}}=|L|^{(1-\frac{2}{n})}$. The worst case w.r.t. $n$ is reached when $m=2$ and $n=\log_2(|L|)$. Therefore, $PoU_{\textnormal{Cor/No}}=\frac{|L|}{4}$.
\end{example}
\begin{example}[Lower bound for worst-case  $PoU_{\textnormal{Com/Cor}}$]\label{ex:comm_cor}
Consider the game presented in Example~\ref{ex:comm_m}. Since $v_{\textnormal{Com}}=1$ and $v_{\textnormal{Cor}}=1/m$, it follows $PoU_{\textnormal{Com/Cor}}=m$. The structure of the game tree is such that $|L|=m^{n-1}$ and thus $PoU_{\textnormal{Com/Cor}}=|L|^{\frac{1}{n-1}}$. Notice that, in this case, the inefficiency is maximized when $n=3$, which corresponds to having a team of two members. Thus, in the worst case w.r.t. $n$, $PoU_{\textnormal{Com/Cor}}=\sqrt{|L|}$.
\end{example}

\section{Finding a TMECom}\label{sec:05_alg_tmecom}

We show that there is a polynomial-time TMECom-finding algorithm. Indeed, we prove that the problem of finding a TMECom is equivalent to finding a 2-player maxmin strategy in an auxiliary 2-player game with perfect recall and that the auxiliary game can be built in polynomial time. 

First, we define the structure of the auxiliary game we use. Let $\Gamma$ be an extensive-form game and $Q=\bigcup_{i\in N} Q_i$. We define the following functions. Function $\mathsf{lead}: V\cup L\rightarrow 2^Q$ returns the sequence profile constituting the path from the root to a given node of the tree. Function $\mathsf{path}:V \times 2^{N} \rightarrow 2^{Q}$ is s.t., for each $x\in V$ and each set of players $G\subseteq N$, 

\begin{scriptsize}
\[
\mathsf{path}(x|G)=\Bigg\{q\subset \bigcup_{i \in G} Q_i \Bigg |\exists q'\subset \bigcup_{i\in N\backslash G}Q_i\wedge  q\cup q'=\mathsf{lead}(x)\Bigg\}.
\]
\end{scriptsize}

\noindent Intuitively, $\mathsf{path}(x|G)$ returns the unique profile of sequences of players in~$G$ leading to $x$ when combined with some sequences of the  players in $N\setminus G$.

The following definition describes the information structure of the auxiliary extensive-form game.

\begin{definition}[$G$-observable game]\label{def:observable_tree}
For any game $\Gamma=(N,A,V,L,\iota,\rho,\chi,H)$ and any set of players $G\subseteq N$, the $G$-observable game $\hat{\Gamma}$ is a tuple $(N,A,V,L,\iota,\rho,\chi,\hat{H})$, where $\hat{H}=\left(\bigcup_{i\in G}\hat{H}_i\right) \cup \left(\bigcup_{i\in N\backslash G} H_i\right)$ is such that:
\begin{enumerate}
  \item for each decision node $x\in V$, there exists one and only one $\hat{h}\in\hat{H}$ s.t. $x\in\hat{h}$ and $\iota(h)=\iota(\hat{h})$ where  $h$ denotes the information set containing $x$ in $\Gamma$;
  \item for each player $i\in G$, $\hat{H}_i$ is 
  the set with the lowest possible cardinality s.t.
  for each $\hat{h}\in\hat{H}_i$ and for each pair of decision nodes $x,x'\in \hat{h}$, it holds: 
  \[
  \Big(\mathsf{path}(x|G)=\mathsf{path}(x'|G)\Big) \wedge \Big(\exists h\in H_i| x\in h\wedge x'\in h\Big).
  \]
\end{enumerate}
\end{definition}
In a $G$-observable extensive-form game, players belonging to $G$ are fully aware of the moves of other players in $G$ and share the same information on the moves taken by players in $N\backslash G$. We show that we can build $\hat{\Gamma}$ in polynomial time.

\begin{lemma}[$T$-observable game construction]\label{lemma:observable_tree}\ \\
The T-observable game $\hat{\Gamma}$ of a generic STSA-EF-TG $\Gamma$ can be computed in polynomial time.
\end{lemma}
\textbf{Proof.} We provide the sketch of an algorithm  (the pseudocode is provided in the Appendices) to  build a $T$-observable game (i.e., a $G$-observable game with $G=T$) in time and space polynomial in the size of the game tree. 
The algorithm employs nested hash-tables. The first hash-table associates each joint sequence of the team with another hash-table, which is indexed over information sets and has as value the information set id to be used in $\hat{\Gamma}$. $\Gamma$ is traversed in a depth-first manner while keeping track of the sequence leading to the current node. For each $x\in V$ s.t. $\iota(x)\in T$, a search/insertion over the first hash-table is performed by hashing $\mathsf{path}(x|T)$. Then, once the sequence-specific hash-table is found, the information set is assigned a new id if it is not already present as a key. $\hat{\Gamma}$ is built by associating to each decision node of the team a new information set as specified in the hash-table. The worst-case running time is $O(|V|^{2})$.\hfill$\Box$

\begin{theorem}[TMECom computation]\label{th:comm_eq_poly}
Given a STSA-EF-TG and a communication device for $\T$, the unique (unless degeneracy) TMECom can be found in polynomial time.
\end{theorem}
\textbf{Proof}. Given a STSA-EF-TG $\Gamma$, the use of a communication device for the team $\T$ changes the information structure of the game inducing a $T$-observable game $\hat{\Gamma}$. A TMECom can be computed over $\hat{\Gamma}$ as follows.
Given a communication device $(H_\T,A_\T,R^{\textnormal{Com}})$, $R^{\textnormal{Com}}$ enforces a probability distribution  $\gamma$ over the set of feedback rules. $\gamma$ is chosen in order to maximize the expected utility of the team. In this setting, no incentive constraints are required because teammates share the same utility function and therefore, under the hypothesis that $\gamma$ maximizes it, it is in their best interest to follow the recommendations sent by the device and to report truthfully their information. Thus, considering the function $\mathsf{path}$ to be defined over information sets and $\hat{H}_\T=\bigcup_{i\in T}\hat{H}_i$, $\gamma$ reduces to a distribution over rules of type $\{\beta=(\beta^{h})_{h\in \hat{H}_\T}| \beta^h:\mathsf{path}(h|T)\rightarrow \rho(h),\forall h\in\hat{H}_\T\}$.

We are left with an optimization problem in which we have to choose $\gamma$ s.t. the worst-case utility of the team is maximized. This is equivalent to a 2-player maxmin problem over $\hat{\Gamma}$ between $\A$ and a player playing over team's joint sequences. By construction, the team player has perfect recall and thus the maxmin problem can be formulated as an LP in sequence form, requiring polynomial time.\hfill$\Box$
\section{Finding a TMECor}\label{sec:05_alg_tmecor}

We initially focus on the computational complexity of the problem of searching for a TMECor.
\begin{theorem}[TMECor complexity]\label{th:nf_cor_hardness}
	Finding a TMECor is \textsf{FNP}-hard when there are two teammates, each with an arbitrary number of information sets, or when there is an arbitrary number of teammates, each with one information set.
\end{theorem}
The first result directly follows from the reduction presented in~\cite[Theorem 1.3]{vonStengel2008} since the game instances used in the reduction are exactly STSA-EF-TGs with 2 teammates. The second  result can be proved by adapting the reduction described in~\cite[Proposition 2.6]{koller1992}, assigning each information set of the game instances to a different teammate.

In principle, a TMECor can be found by casting the game in normal form and then by searching for a Team-maxmin equilibrium with correlated strategies. This latter equilibrium can be found in polynomial time in the size of the normal form, which, however, is given by $P_1\times \ldots \times P_n$, where each $P_i$ is exponentially large in the size of the tree. We provide here a more efficient method that can also be used in an anytime fashion, without requiring any exponential enumeration before the execution of the algorithm. In our method, we use a hybrid representation that, to the best of our knowledge, has not been used in previous works.

\textbf{Hybrid representation}. In our representation, $\mathcal{A}$'s strategy is represented in sequence form, while the team plays over \emph{jointly-reduced plans}, as formally defined below.  Given a generic STSA-EF-TG $\Gamma$, let us denote with $P_r=\{P_{r,1},\ldots,P_{r,n}\}$ the set of actions of the reduced normal-form of $\Gamma$, where $P_{r,i}$ is the set of reduced plans for player $i$. Therefore, $\bigtimes_{i\in T} P_{r,i}$ is the set of joint reduced plans of the team. Let function $\mathsf{terminal}: Q_\A\times \{\bigtimes_{i\in T} P_{r,i}\} \rightarrow L\cup\{\varnothing\}$ be s.t. it returns, for a given pair $(q_\A, p)$, the terminal node reached when the adversary plays $q_\A$ and the team members, at each of their information set, play according to $p$. If no terminal node is reached, $\varnothing$ is returned. We define some equivalence classes over $\bigtimes_{i\in T} P_{r,i}$ by the relation $\sim$:

\begin{definition}\label{def:eq_rel}
	The equivalence relation $\sim$ over $\bigtimes_{i\in T} P_{r,i}$ is s.t., given $p_1, p_2\in \bigtimes_{i\in T} P_{r,i}$, $p_1\sim p_2$ iff, for each $q_\A \in Q_\A$, $\mathsf{terminal(q_\A, p_1)}=\mathsf{terminal(q_\A,p_2)}$.
\end{definition} 

\begin{definition}[Jointly-reduced plans]\label{def:joint_red}
	The set of jointly-reduced plans $P_{jr}\subseteq \bigtimes_{i\in T} P_{r,i}$ is obtained by picking exactly one representative from each equivalence class of $\sim$.
\end{definition}

\noindent The team's utility function is represented by the sparse $|Q_\A|\times |P_{jr}|$ matrix $U_h$. Given a pair $(q_\mathcal{A}, p_{jr})\in Q_\A\times P_{jr}$,  $U_\T(\mathsf{terminal}(q_\mathcal{A}, p_{jr}))$ is stored in $U_h$ iff $\mathsf{terminal}(q_\mathcal{A}, p_{jr})\neq\varnothing$. Notice that $U_h$ is well defined since each pair $(q_\mathcal{A}, p_{jr})$ leads to at most one terminal-node.

Let $\sigma_\T$ denote the team's strategy over $P_{jr}$. The problem of finding a TMECor in our hybrid representation can be formulated as the following LP named \textsc{hybrid-maxmin}:

\vspace{-0.2cm}
\begin{scriptsize}
\begin{align*}
	\argmax_{\sigma_\T,v} & \hspace{0.5cm} \sum_{h\in H_\mathcal{A}\cup\{h_\emptyset\}}f_\mathcal{A}(h)v(h) \hspace{0.5cm} \text{s.t.} \\
	&\sum_{h\in H_\mathcal{A}\cup\{h_\emptyset\}} \hspace{-0.5cm} F_\mathcal{A}(h,q_\mathcal{A})v(h) - \hspace{-0.2cm} \sum_{p\in P_{jr}} \hspace{-0.15cm} U_h(q_\mathcal{A},p)\sigma_\T(p)\leq 0  \hspace{0.2cm} \forall q_\mathcal{A}\in Q_\mathcal{A} \\
	&\sum_{p\in P_{jr}}\sigma_\T(p)=1 & \quad \\
	&\sigma_\T(p)\geq 0 \hspace{5.25cm} \forall p\in P_{jr}
\end{align*}
\end{scriptsize}
\vspace{-0.2cm}

\noindent composed of $|Q_\mathcal{A}|+1$ constraints (except $\sigma_\T(p)\geq 0$ constraints) and an exponential number of variables $\sigma_\T$. Thus, we can state the following proposition.

\begin{prop}\label{prop:strat}
	There exists at least one TMECor in which the number of joint plans played with strictly positive probability by the team is at most $|Q_\mathcal{A}|$.
\end{prop}

\noindent \textbf{Proof}.\label{proof:strat}
	The above LP admits a basic optimal solution with at most $|Q_\mathcal{A}|+1$ variables with strictly positive values~\cite{shapley1950}. Since $v$ is always in the basis (indeed, we can add a constant to make the team's utility in each terminal node strictly positive without affecting equilibrium strategies), the joint plans in the basis are $|Q_\mathcal{A}|$.\hfill$\Box$

Proposition~\ref{prop:strat} shows that the \textsf{NP}-hardness of the problem is merely due to guessing the jointly-reduced plans played with strictly positive probability in a TMECor. Thus, we can avoid enumerating entirely $P_{jr}$ before executing the algorithm by working with a subset of jointly-reduced plans built progressively, in a classical column-generation fashion (see, e.g.,~\cite{mcmahan2003}).

\textbf{Column-generation algorithm}. The pseudocode is given in Algorithm~\ref{alg:column_gen}. It receives in input the game tree and the sequence-form constraint matrices $F_i$ of all the players (Line~\ref{alg:input}). Then, the algorithm is initialized, assigning a matrix of zeros to $U_h$, an empty set to $P_{cur}$, and 0 to $v$ (Line~\ref{alg:init_start}). Notice that $U_h$ is sparse and therefore its representation requires a space equal to the number of non-null entries. $\overline{r}_\A$ is initialized as a realization plan equivalent to a uniform behavioral mixed strategy, i.e., the adversary, at each information set, randomizes uniformly over all the available actions (Line~\ref{alg:init_ra}). Then, the algorithm calls the \textsf{BR-ORACLE} (defined below) to find the best response of the team given the adversary's strategy $\overline{r}_\A$ (Line~\ref{alg:first_oracle_call}). Lines~\ref{alg:update}-\ref{alg:oracle_call} are repeated until an optimal solution is found. Initially, $br$ is added to $P_{cur}$ (Line~\ref{alg:update}) and players' utilities at nodes reached by $(q_\A,br)$ for every $q_\A$ are added to $U_h$. Then, the algorithm solves the maxmin (\textsc{hybrid-maxmin}) and minmax (\textsc{hybrid-minmax}) problems restricted to $P_{cur}$ (Lines~\ref{alg:maxmin} and \ref{alg:minmax}), where the \textsc{hybrid-minmax} problem is defined as:

\begin{scriptsize}
		\begin{align*}
		\argmin_{r_\A,v} & \hspace{0.5cm} v\hspace{0.5cm} \text{s.t.}\\
		& v - \sum_{q\in Q_\A}U_h(q,p_{jr})r_\A(q) \geq 0 \hspace{1.5cm} \forall p_{jr}\in P_{jr}\\
		&\sum_{q\in Q_\A}F_\A(h,q)=f_\A(h)\hspace{2.5cm}\forall h\in H_\A\\
		&r_\A(q)\geq 0\hspace{4.0cm}\forall q\in Q_\A
		\end{align*}
\end{scriptsize}

\noindent Finally, the algorithm calls \textsf{BR-ORACLE} to find the best response to $\overline{r}_\A$ (Line~\ref{alg:oracle_call}).

\textbf{Best-response oracle.} Given a generic STSA-EF-TG $\Gamma$, we denote the problem of finding the best response of the team against a given a fixed realization plan $\overline{r}_\A$ of the adversary over $\Gamma$ as \textsf{BR-T}. This problem is shown to be \textsf{NP}-hard in the reduction used for~\cite[Theorem 1.3]{vonStengel2008}, where we can interpret the initial chance move as the fixed strategy of the adversary. We can strengthen such a hardness result as follows (the proofs are provided in the Appendices):

\begin{theorem}\label{th:br_apx_hard}
	\textsf{BR-T} is \textsf{APX}-hard.
\end{theorem}

Let $\alpha_{(\cdot)}\in[0,1]$ be the best approximation bound of the maximization problem $(\cdot)$. 

\begin{theorem}\label{th:br_apx_bound}
	Denote with $\textsf{BR-T-h}$ the problem \textsf{BR-T} over STSA-EF-TG instances of fixed maximum depth $3h$ and branching factor variable at each decision-node, it holds:
	$$\alpha_{\textsf{BT-T-h}}\leq(\alpha_{\textsf{MAX-SAT}})^h.$$
\end{theorem}
This means that the upper bound on the approximation factor decreases exponentially as the depth of the tree increases\footnote{Notice that $\alpha_{\textsf{MAX-SAT}}\leq 7/8$, see~\cite{haastad2001}.}.
The column-generation oracle solving \textsf{BR-T} can be formulated as the following integer linear program (ILP):

\begin{scriptsize}
	\begin{align*}
	\argmax_{r_1,\ldots,r_{(n-1)},x} & \hspace{0.5cm} \sum_{l\in L}U_\T(l)x(l)\overline{r}_\mathcal{A}(\mathsf{path}(l|\{n\})) \hspace{0.5cm} \text{s.t.} \\
	& \sum_{q_i\in Q_i} F_i(h,q_i)r_i(q_i)=f_i(h) & \substack{\forall i\in T,\\ \forall h\in H_i\cup\{h_\emptyset\}} \\
	& x(l)\leq r_i(q_i) & \substack{\forall i\in T, \forall l\in L,\\ \forall q_i\in \mathsf{path}(l|\{i\})} \\
	& x(l)\in \{0,1\} & \forall l\in L
	\end{align*}
\end{scriptsize}

\noindent where $x(l)$ is a binary variable which is equal to $1$ iff, for all the sequences $q_i\in Q$ necessary to reach $l$, it holds $r_i(q_i)=1$. Notice that the oracle returns a pure realization plan for each of the teammates. Team's best-response is a jointly-reduced realization plan that can be derived as follows. Denote with $Q^{L}_i$ the set of sequences played with probability one by $i$ that are not subsets of any other sequence played with positive probability. Let $p'_{i}$ be the reduced normal-form plan of player $i$ specifying all and only actions played in the sequences belonging to $Q^L_i$. The joint plan $p'=(p'_1,\ldots,p'_{n-1})$ is s.t. $p'\in P_{jr}$.

\begin{algorithm}[!t]
	\caption{\texttt{Hybrid Column Generation}}
	\begin{scriptsize}
		\begin{algorithmic}[1]
			\Function{HYBRID-COL-GEN}{$\Gamma$, $F_1, \ldots, F_{n-1},F_\A$}\Comment{$\Gamma$ is a generic STSA-EF-TG and $F_i$ are sequence-form constraint matrices}\label{alg:input}
			\State $U_h = \mathbf{0}$, $P_{cur}=\{\}$, $v\leftarrow 0$ \Comment{initialization}\label{alg:init_start}	
			\State $\overline{r}_\A \leftarrow$ realization plan equivalent to a uniform  behavioral mixed strategy \label{alg:init_ra}
			\State $br\leftarrow \textsf{BR-ORACLE}(\Gamma, \{F_1,\ldots,F_{n-1}\},\overline{r}_\A)$\Comment{call to the oracle}\label{alg:first_oracle_call}
			\While{$br\notin P_{cur}$}
				\State $P_{cur}\leftarrow$ $P_{cur} \cup br$\label{alg:update}
				\State players' utilities in $(q_\A,br)$ for every $q_\A$ are added to $U_h$ \label{alg:update}
				\State $\sigma_{\mathcal{T}}\leftarrow$ solve \textsc{hybrid-maxmin} problem with $(U_h, P_{cur}, F_\A)$\label{alg:maxmin}
				\State $\overline{r}_\A\leftarrow$ solve \textsc{hybrid-minmax} problem with $(U_h, P_{cur}, F_\A)$\label{alg:minmax}
				\State $br\leftarrow \textsf{BR-ORACLE}(\Gamma, \{F_1,\ldots,F_{n-1}\},\overline{r}_\A)$\label{alg:oracle_call}
			\EndWhile
			\State \textbf{return} $(\overline{r}_{\mathcal{A}},\sigma_{\mathcal{T}})$
			\EndFunction
		\end{algorithmic}
	\end{scriptsize}
	\label{alg:column_gen}
\end{algorithm}

A simple approximation algorithm can be obtained by a continuous relaxation of the binary constraints $x(l)\in \{0,1\}$. The resulting mathematical program is linear and therefore solvable in polynomial time. An approximated solution can be obtained by randomized rounding~\cite{raghavan1987}. When considering game trees encoding \textsf{MAX-SAT} instances (see the proof of Theorems~\ref{th:br_apx_hard}), the approximation algorithm matches the ratio guaranteed by randomized-rounding for \textsf{MAX-SAT} (details are given in the Appendices).

\section{Finding a TME}\label{sec:06_alg_tme}

We recall that finding a TME is hard, since it is hard even with normal-form games~\cite{hansen2008}.
\begin{theorem}[TME complexity]
 	Finding a TME is \textsf{FNP}-hard and its value is inapproximable in additive sense even with binary payoffs. 
\end{theorem}
The problem of finding a TME can be formulated as the following non-linear mathematical programming problem:

\vspace{-0.1cm}
\begin{scriptsize}
\begin{align*}
\max_{r_1,\ldots,r_{(n-1)}} & \hspace{0.5cm} v(h_\emptyset) \hspace{0.5cm} \text{s.t.}\\
&\sum_{h\in H_\mathcal{A}\cup\{h_\emptyset\}}F_\mathcal{A}(h,q_\mathcal{A})v(h) \leq\\
& \hspace{0.5cm} \leq \sum_{q_\T\in Q_\T}(U_\T(q_\T,q_\mathcal{A})\prod_{i\in T}r_i(q_\T(i))) & \quad \forall q_\mathcal{A}\in Q_\mathcal{A} \\
& \sum_{q_i\in Q_i}F_i(h,q_i)r_i(q_i)=f_i(h) & \quad \substack{\forall i\in T,\\\forall h\in H_i\cup\{h_\emptyset\}} \\
& r_{i}(q_i)\geq 0 & \quad \substack{\forall i\in T,\\ \forall q_i\in Q_i}\\
\end{align*}
\end{scriptsize}
\vspace{-0.6cm}

\noindent where $Q_\T$ is the set of team's joint sequences and $q_\T(i)$ identifies the sequence of player $i$ in $q_\T$. This program can be solved exactly, within a given numerical accuracy, by means of global optimization tools in exponential time.

\section{Experimental Evaluation}\label{sec:exp}

\begin{figure*}[t]
\centering
\includegraphics[width=0.25\textwidth]{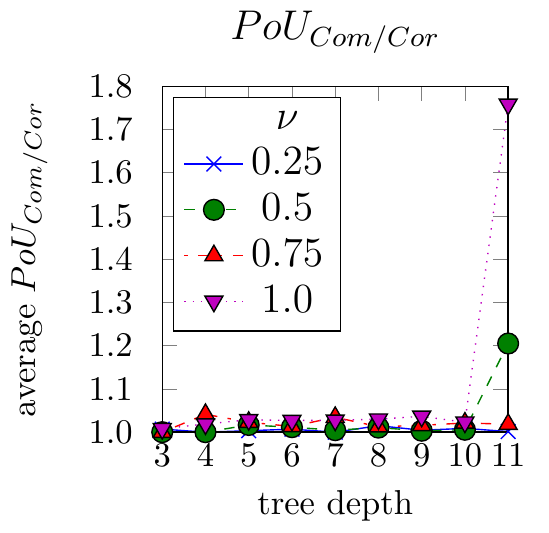}
\includegraphics[width=0.25\textwidth]{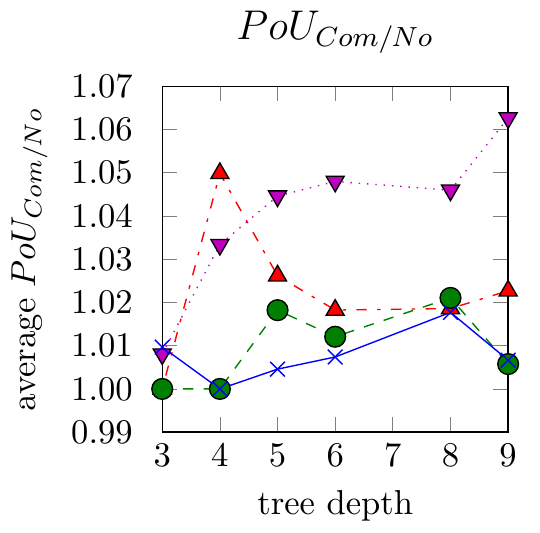}
\includegraphics[width=0.25\textwidth]{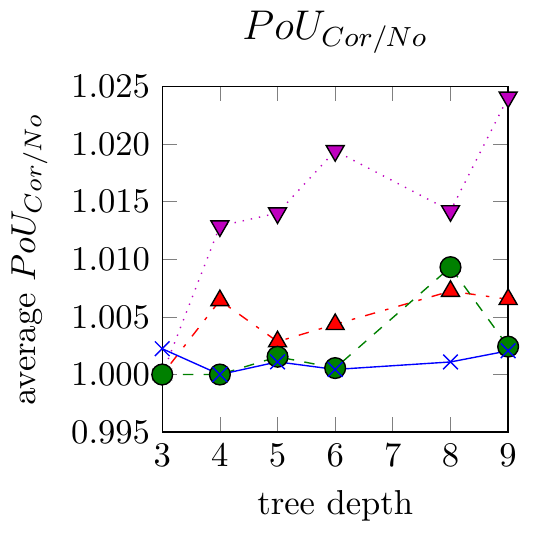}
\caption{Average empiric inefficiency indices with 3 players and some values of $\nu$.}
\label{fig:empiricindices}
\end{figure*}

\begin{figure*}[t]
\centering
\includegraphics[width=0.72\textwidth]{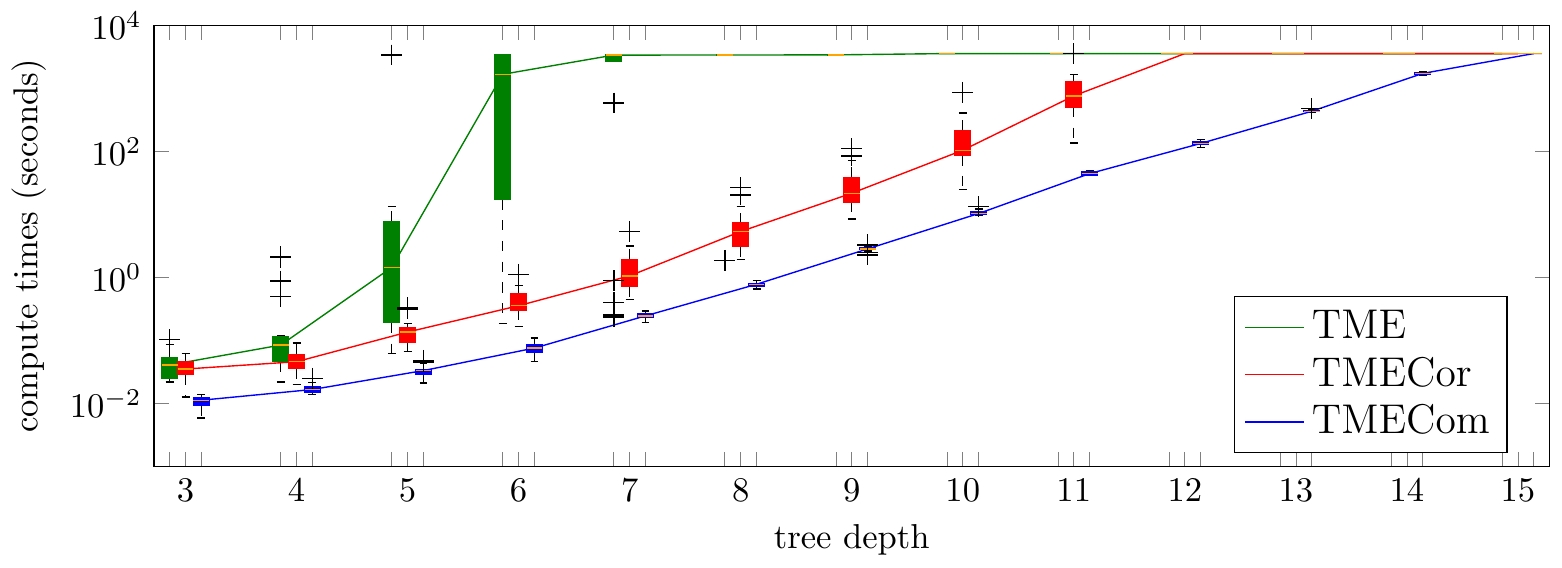}
\caption{Average compute times of the algorithms and their box plots with 3 players and $\nu=0.5$.}
\label{fig:computetimes}
\end{figure*}

\textbf{Experimental setting.} Our experimental setting is based on randomly generated STSA-EF-TGs. The random game generator takes as inputs: the number $n$ of players, a probability distribution over the number of actions available at each information set, the maximum depth $d$ of the tree, and a parameter $\nu$ for tuning the information structure of the tree. Specifically, this parameter encodes the probability with which a newly created decision-node, once it has been randomly assigned to a player, is assigned to an existing information-set (thus, when it is equal to 0 the game is with perfect information), while guaranteeing perfect recall for every player. Finally, payoffs associated with terminal nodes are randomly drawn from a uniform distribution over $[0,1]$. We generate 20 game instances for each combination of the following parameters' values: $n\in\{3,4,5\}$, $d\in\{n,\ldots,15\}$ with step size 1 (i.e., for games with 5 players, $d\in\{5,6,\ldots,15\}$), $\nu\in\{0.0, 0.25, 0.5, 0.75, 1.0\}$. For simplicity, we fix the branching factor to 2 (this value allows us to maximize $d$ and it is also the worst case for the inefficiency indices).

The algorithms are implemented in Python 2.7.6, adopting GUROBI 7.0 for LPs and ILPs, AMPL 20170207 and global optimization solver BARON 17.1.2~\cite{tawarmalani2005}. We set a time limit to the algorithms of 60 minutes. All the algorithms are executed on a UNIX computer with 2.33GHz CPU and 128 GB RAM. We discuss the main experimental results with 3 players below, while the results with more players are provided in the Appendices. Since the computation of the TMECor from the reduced normal form is impractical for $d\geq5$ (see the Appendices), we use only Algorithm~\ref{alg:column_gen} employing the exact oracle (this demonstrated very fast on every instance).

\textbf{Empirical PoUs.} We report in Fig.~\ref{fig:empiricindices} the average empiric inefficiency indices with 3 players for some values of $\nu$. We observe that, despite the theoretical worst-case value increases in $L$, the empiric increase, if any, is negligible. For instance, the worst-case value of $PoU_{Com/Cor}$ with $n=3$ and $L=2^{11}$  is $>45$, while the average empiric value is $<2$. We also observe that the inefficiency increases in $\nu$, suggesting that it may be maximized in normal-form games.

\textbf{Compute time.} We report in Fig.~\ref{fig:computetimes} the average compute times of the algorithms and their box plots with 3 players and $\nu=0.5$ (the plot includes instances reaching the time limit as this not affects results presentation). As expected, the TMECom computation scales well, allowing one to solve games with more than 16,000 terminal nodes in the time limit.  The performances of Algorithm~\ref{alg:column_gen} (TMECor) are remarkable since it solves games with more than 2,000 terminals in the time limit, and presents a narrow boxplot, meaning that the variance in the compute time is small. Notice that, with $d\leq 10$, the compute times of TMECom and TMECor are comparable, even if the former is computationally hard while the latter is solvable in polynomial-time. As expected, the TME computation does not scale well and its compute time is extremely variable among different instances.


\vspace{-0.15cm}
\section{Conclusions}

In this paper, we focus on extensive-form team games with a single adversary. Our main contributions include the definition of game models employing different correlation devices and their suitable solution concepts. We study the inefficiency a team incurs employing various forms of correlation, providing lower bounds to the worst-case values of the inefficiency indices that are arbitrarily large in the game tree.  Furthermore, we study the complexity of finding the equilibria, and we provide exact algorithms. Finally, we experimentally evaluate the scalability of our algorithms and the empirical equilibrium inefficiency in random games. 
In the future, it would be interesting to study approximate equilibrium-finding algorithms in order to reach an improved scalability in all the three correlation scenarios.

\FloatBarrier

\clearpage
\bibliographystyle{aaai}
\bibliography{aaai18}

\begin{thebibliography}{}

\bibitem[\protect\citeauthoryear{Aumann}{1974}]{aumann1974}
Aumann, R.
\newblock 1974.
\newblock Subjectivity and correlation in randomized strategies.
\newblock {\em Journal of Mathematical Economics} 1(1):67--96.

\bibitem[\protect\citeauthoryear{Ausiello, Crescenzi, and
  Protasi}{1995}]{ausiello1995}
Ausiello, G.; Crescenzi, P.; and Protasi, M.
\newblock 1995.
\newblock Approximate solution of np optimization problems.
\newblock {\em Theoretical Computer Science} 150(1):1--55.

\bibitem[\protect\citeauthoryear{Basilico \bgroup et al\mbox.\egroup
  }{2017}]{basilico2016}
Basilico, N.; Celli, A.; Nittis, G.~D.; and Gatti, N.
\newblock 2017.
\newblock Team-maxmin equilibrium: efficiency bounds and algorithms.
\newblock In {\em AAAI}.

\bibitem[\protect\citeauthoryear{Borgs \bgroup et al\mbox.\egroup
  }{2010}]{borgs2010}
Borgs, C.; Chayes, J.~T.; Immorlica, N.; Kalai, A.~T.; Mirrokni, V.~S.; and
  Papadimitriou, C.~H.
\newblock 2010.
\newblock The myth of the folk theorem.
\newblock {\em Games and Economic Behavior} 70(1):34--43.

\bibitem[\protect\citeauthoryear{Conitzer and
  Sandholm}{2006}]{conitzer2006computing}
Conitzer, V., and Sandholm, T.
\newblock 2006.
\newblock Computing the optimal strategy to commit to.
\newblock In {\em ACM EC},  82--90.

\bibitem[\protect\citeauthoryear{Forges}{1986}]{forges1986}
Forges, F.
\newblock 1986.
\newblock An approach to communication equilibria.
\newblock {\em Econometrica}  1375--1385.

\bibitem[\protect\citeauthoryear{Gatti \bgroup et al\mbox.\egroup
  }{2012}]{gatti2012}
Gatti, N.; Patrini, G.; Rocco, M.; and Sandholm, T.
\newblock 2012.
\newblock Combining local search techniques and path following for bimatrix
  games.
\newblock In {\em UAI},  286--295.

\bibitem[\protect\citeauthoryear{Hansen \bgroup et al\mbox.\egroup
  }{2008}]{hansen2008}
Hansen, K.~A.; Hansen, T.~D.; Miltersen, P.~B.; and S{\o}rensen, T.~B.
\newblock 2008.
\newblock Approximability and parameterized complexity of minmax values.
\newblock In {\em {WINE}},  684--695.

\bibitem[\protect\citeauthoryear{H{\aa}stad}{2001}]{haastad2001}
H{\aa}stad, J.
\newblock 2001.
\newblock Some optimal inapproximability results.
\newblock {\em Journal of the ACM (JACM)} 48(4):798--859.

\bibitem[\protect\citeauthoryear{Kohlberg and Mertens}{1986}]{kohlberg1986}
Kohlberg, E., and Mertens, J.-F.
\newblock 1986.
\newblock On the strategic stability of equilibria.
\newblock {\em Econometrica: Journal of the Econometric Society}  1003--1037.

\bibitem[\protect\citeauthoryear{Koller and Megiddo}{1992}]{koller1992}
Koller, D., and Megiddo, N.
\newblock 1992.
\newblock The complexity of two-person zero-sum games in extensive form.
\newblock {\em Games and economic behavior} 4(4):528--552.

\bibitem[\protect\citeauthoryear{Kuhn}{1953}]{kuhn1953}
Kuhn, H.~W.
\newblock 1953.
\newblock {\em Extensive Games and the Problem of Information}.
\newblock Princeton University Press.
\newblock  193--216.

\bibitem[\protect\citeauthoryear{Lemke and {Howson,
  Jr}}{1964}]{lemke64equilibrium}
Lemke, C.~E., and {Howson, Jr}.
\newblock 1964.
\newblock {Equilibrium Points of Bimatrix Games}.
\newblock {\em Journal of the Society for Industrial and Applied Mathematics}
  12(2):413--423.

\bibitem[\protect\citeauthoryear{McMahan, Gordon, and Blum}{2003}]{mcmahan2003}
McMahan, H.~B.; Gordon, G.~J.; and Blum, A.
\newblock 2003.
\newblock Planning in the presence of cost functions controlled by an
  adversary.
\newblock In {\em Proceedings of the 20th International Conference on Machine
  Learning (ICML-03)},  536--543.

\bibitem[\protect\citeauthoryear{Nisan \bgroup et al\mbox.\egroup
  }{2007}]{nisan2007algorithmic}
Nisan, N.; Roughgarden, T.; Tardos, E.; and Vazirani, V.
\newblock 2007.
\newblock {\em Algorithmic game theory}, volume~1.
\newblock Cambridge University Press.

\bibitem[\protect\citeauthoryear{Raghavan and Tompson}{1987}]{raghavan1987}
Raghavan, P., and Tompson, C.~D.
\newblock 1987.
\newblock Randomized rounding: a technique for provably good algorithms and
  algorithmic proofs.
\newblock {\em Combinatorica} 7(4):365--374.

\bibitem[\protect\citeauthoryear{Selten}{1975}]{selten1975}
Selten, R.
\newblock 1975.
\newblock Reexamination of the perfectness concept for equilibrium points in
  extensive games.
\newblock {\em International journal of game theory} 4(1):25--55.

\bibitem[\protect\citeauthoryear{Shapley and Snow}{1950}]{shapley1950}
Shapley, L.~S., and Snow, R.~N.
\newblock 1950.
\newblock Basic solutions of discrete games.
\newblock {\em Annals of Mathematics Studies} 24:27--35.

\bibitem[\protect\citeauthoryear{Shoham and
  Leyton-Brown}{2009}]{shoham2009multiagent}
Shoham, Y., and Leyton-Brown, K.
\newblock 2009.
\newblock Multiagent systems: Algorithmic, game-theoretic, and logical
  foundations.

\bibitem[\protect\citeauthoryear{Tambe}{2011}]{tambe2011}
Tambe, M.
\newblock 2011.
\newblock {\em Security and game theory: algorithms, deployed systems, lessons
  learned}.
\newblock Cambridge University Press.

\bibitem[\protect\citeauthoryear{Tawarmalani and
  Sahinidis}{2005}]{tawarmalani2005}
Tawarmalani, M., and Sahinidis, N.~V.
\newblock 2005.
\newblock A polyhedral branch-and-cut approach to global optimization.
\newblock {\em Mathematical Programming} 103:225--249.

\bibitem[\protect\citeauthoryear{von Stengel and Forges}{2008}]{vonStengel2008}
von Stengel, B., and Forges, F.
\newblock 2008.
\newblock Extensive-form correlated equilibrium: Definition and computational
  complexity.
\newblock {\em Mathematics of Operations Research} 33(4):1002--1022.

\bibitem[\protect\citeauthoryear{von Stengel and
  Koller}{1997}]{vonStengelKoller1997}
von Stengel, B., and Koller, D.
\newblock 1997.
\newblock Team-maxmin equilibria.
\newblock {\em Games and Economic Behavior} 21(1):309 -- 321.

\bibitem[\protect\citeauthoryear{von Stengel}{1996}]{vonStengel1996}
von Stengel, B.
\newblock 1996.
\newblock Efficient computation of behavior strategies.
\newblock {\em Games and Economic Behavior} 14(2):220 -- 246.

\bibitem[\protect\citeauthoryear{Williamson and Shmoys}{2011}]{williamson2011}
Williamson, D.~P., and Shmoys, D.~B.
\newblock 2011.
\newblock {\em The design of approximation algorithms}.
\newblock Cambridge university press.

\end{thebibliography}

\clearpage
\appendix
\section*{Appendices}
\section{Proofs of the Theorems}\label{sec:finding_tmecor}

\setcounter{theorem}{3}

\begin{theorem}\label{th:br_apx_hard}
	\textsf{BR-T} is \textsf{APX}-hard.
\end{theorem}
\noindent \textbf{Proof}.	We prove that \textsf{MAX-SAT} is AP-reducible to \textsf{BR-T} (\textsf{MAX-SAT}$\leq_{AP}$\textsf{BR-T}). Given a boolean formula $\phi$ in conjunctive normal form, \textsf{MAX-SAT} is the problem of determining the maximum number of clauses that can be made true by a truth assignment to variables of $\phi$. 
	
For any $\phi$ with $c$ clauses, we build, with a construction similar to~\cite[Theorem 1.3]{vonStengel2008}, a STSA-EF-TG $\Gamma$ as follows: 
	\begin{itemize}
		\item $N=\{\A, T\}$, and $T=\{1,2\}$;
		\item $\A$ plays first and has a unique decision node $x_\A$ (the root of the tree), s.t. $|\rho(x_\A)|=c$;
		\item player 1 plays on the second level of the tree and has a singleton information set for each clause in $\phi$. Each information set has, as its actions, the variables that appear in the clause it identifies;
		\item player 2 plays on the third level of the tree. She has one information set for each literal of $\phi$. At each of her information sets, player 2 chooses whether the literal has to be positive or negative;
		\item $U_\T=1$ if the literal chosen by player 1 is true in the assignment made by player 2.
	\end{itemize}
	Consider $\A$ to be randomizing uniformly over her actions. With this construction, $\Gamma$ has a pair of pure strategies for the team members leading to payoff 1 iff $\phi$ is satisfiable. Let $f(\phi)$ denote the extensive-form game $\Gamma_\phi$ obtained by the above construction starting from $\phi$. Denote with $br$ the solution to \textsf{BR-T} for $f(\phi)$. Function $g(\phi,\Gamma_\phi, br)$ maps the best-response result back to a feasible assignment for the \textsf{MAX-SAT} problem.  
	
	Once fixed $\overline{r}_\A$ so that each terminal sequence of $\A$ is selected with probability $1/c$, the objective functions of \textsf{MAX-SAT} and \textsf{BR-T} are equivalent since maximizing the utility of the team implies finding the maximum number of satisfiable instances in $\phi$. Denote with $\mathsf{OBJ}(\Gamma)_{BR}$ and $\mathsf{OBJ}(\phi)_{MS}$ the value of the two objective functions of \textsf{BR-T} and  \textsf{MAX-SAT}, respectively. It holds $\frac{1}{c}\mathsf{OBJ}(\phi)_{MS}=\mathsf{OBJ}(f(\phi))_{BR}$. For this reason, the AP-condition holds. Specifically, for any $\phi$, for any rational $\alpha>1$, for any feasible solution $br$ to \textsf{BR-T} over $\Gamma_\phi=f(\phi)$, it holds:
	$$\frac{\mathsf{OPT}_{BR}(\Gamma_\phi)}{\mathsf{OBJ}_{BR}(br)}\leq \alpha \implies \frac{\mathsf{OPT}_{MS}(\phi)}{\mathsf{OBJ}_{MS}(g(\phi,\Gamma_\phi,br))}\leq 1+\beta(\alpha-1)$$
	\noindent where $\mathsf{OPT}_{BR}(\cdot)$ and $\mathsf{OPT}_{MS}(\cdot)$ are, respectively, the optimal solutions to a given instance of the two problems and $\beta=1$.
	Therefore, since \textsf{MAX-SAT} is an \textsf{APX}-complete problem (see~\cite{ausiello1995}) and it is AP-reducible to \textsf{BR-T}, \textsf{BR-T} is \textsf{APX}-hard.\hfill$\Box$

\vspace{2cm}

\begin{theorem}
	Denote with \textsf{BR-T-h} the problem \textsf{BR-T} over STSA-EF-TG instances of fixed maximum depth $3h$ and branching factor variable at each decision-node. It holds:
	$$\alpha_{\textsf{BT-T-h}}\leq(\alpha_{\textsf{MAX-SAT}})^h.$$
\end{theorem}
\noindent \textbf{Proof}.
	We recall that $\alpha_{(\cdot)}\in[0,1]$ denotes the best upper-bound for the efficient approximation of maximization problem $(\cdot)$. 
	
	Let $\phi$ be a boolean formula in conjunctive normal form. Fix the maximum depth of the tree to an arbitrary value $3h\geq 1$. Build a STSA-EF-TG $\Gamma_\phi$ following the construction explained in the proof of Theorem~\ref{th:br_apx_hard}. At this point, for each terminal node $x_j\in L$ of $\Gamma_\phi$ s.t. $U_\T(x_j)=1$, replicate $\Gamma_\phi$ by substituting $x_j$ with the root of a new $\Gamma^{x_j}_\phi=\Gamma_\phi$. Repeat this procedure on the terminal nodes of the newly added subtrees until the longest path from the root of $\Gamma_\phi$ to one of the new leafs traverses $h$ copies of the original tree. Denote the full tree obtained through this process with $\Gamma'_\phi$. The maximum depth of $\Gamma'_\phi$ is $3h$ and it contains the set of $\{\Gamma^{x_1}_\phi,\ldots,\Gamma^{x_k}_\phi\}$ replicas of $\Gamma_\phi$. 
	
	Suppose, by contradiction, there exists a polynomial-time approximation algorithm for \textsf{BR-T-h} guaranteeing a constant approximation factor $\alpha'_{\textsf{BR-T-h}}> (\alpha_{\textsf{MAX-SAT}})^h$. Apply this algorithm to find an approximate solution to \textsf{BR-T-h} over $\Gamma'_\phi$. 
	For at least one of the sub-trees in $\{\Gamma_\phi,\Gamma^{x_1}_\phi,\ldots,\Gamma^{x_k}_\phi\}$, it has to hold: $\alpha^{x_j}_{\textsf{BR-T-h}}>\alpha_{\textsf{MAX-SAT}}$, where $\alpha^{x_j}_{\textsf{BR-T-h}}$ is the approximation ratio obtained by the algorithm for the problem \textsf{BR-T-h} over $\Gamma^{x_j}_\phi$. As shown in the proof of Theorem~\ref{th:br_apx_hard}, a solution to \textsf{BR-T} over a tree obtained with our construction can be mapped back to obtain an approximate solution to \textsf{MAX-SAT}. The same reasoning holds for \textsf{BR-T-h}. Therefore, if $\alpha^{x_j}_{\textsf{BR-T-h}}>\alpha_{\textsf{MAX-SAT}}$, then $\alpha^{x_j}_{\textsf{MAX-SAT}}>\alpha_{\textsf{MAX-SAT}}$, where $\alpha^{x_j}_{\textsf{MAX-SAT}}$ is the approximation ratio obtained approximating the \textsf{MAX-SAT} instance by mapping the solution of \textsf{BR-T-h} over $\Gamma^{x_j}_\phi$. Therefore, the approximation algorithm guarantees a constant approximation factor for \textsf{MAX-SAT} which is strictly greater that its theoretical upper bound, which is a contradiction. \hfill$\Box$	

\section{On the approximation algorithm} As mentioned in the main paper, a simple approximation algorithm for the \textsf{BR-T} problem can be obtained by relaxing the binary constraints $x(l)\in \{0,1\}$ and then applying randomized rounding~\cite{raghavan1987}. The linear program relaxation of the ILP oracle is:

\begin{scriptsize}
	\begin{align*}
	\argmax_{r_1,\ldots,r_{(n-1)},x} & \hspace{0.5cm} \sum_{l\in L}U_\T(l)x(l)\overline{r}_\mathcal{A}(\mathsf{path}(l|\{n\})) \hspace{0.5cm} \text{s.t.} \\
	& \sum_{q_i\in Q_i} F_i(h,q_i)r_i(q_i)=f_i(h) & \substack{\forall i\in T,\\ \forall h\in H_i\cup\{h_\emptyset\}} \\
	& x(l)\leq r_i(q_i) & \substack{\forall i\in T, \forall l\in L,\\ \forall q_i\in \mathsf{path}(l|\{i\})} \\
	& 0\leq x(l)\leq 1 & \forall l\in L
	\end{align*}
\end{scriptsize}

Let $(r^*_1,\ldots,r^*_{(n-1)},x^*)$ be an optimal solution to the LP relaxation. We select the approximate best-response, which has to be a pure realization plan for each player, by selecting actions according to probabilities specified by $r^*_1,\ldots,r^*_{(n-1)}$. Notice that, once an action has been selected, probability values at the next decision-node of the team have to be rescaled so that they sum to one (therefore, the rounding process starts from the root). 

Let us focus on games encoding \textsf{MAX-SAT} instances. Specifically, denote with $\phi$ a generic boolean formula in conjunctive normal form and with $\Gamma_\phi$ the STSA-EF-TG built as specified in the proof of Theorem~\ref{th:br_apx_hard}. It is interesting to notice that, for any $\phi$, the application of our approximation algorithm to $\Gamma_\phi$ guarantees the same approximation ratio of randomized rounding applied to the relaxation of the ILP formulation of \textsf{MAX-SAT}. 

Denote with $\mathcal{A}^R_{\textsf{BR-T}}$ and $\mathcal{A}^R_{\textsf{MAX-SAT}}$ the approximate algorithms based on randomized rounding for \textsf{BR-T} and \textsf{MAX-SAT} respectively. The following result holds:
\begin{prop}
	For any $\phi$, the approximation ratio for \textsf{MAX-SAT} over $\phi$ obtained by the solution of \textsf{BR-T} over $\Gamma_\phi$ trough $\mathcal{A}^R_{\textsf{BR-T}}$ is guaranteed to be at least $(1-1/e)$, i.e., the ratio guaranteed by $\mathcal{A}^R_{\textsf{MAX-SAT}}$.
\end{prop}
\textbf{Proof}.
	The relaxation of the \textsf{MAX-SAT} ILP ($\mathcal{A}^R_{\textsf{MAX-SAT}}$) is the following linear formulation:
	\begin{scriptsize}
		\begin{align*}
		\max& \hspace{0.5cm} \sum_{c\in C_\phi} v_c \hspace{0.5cm} \text{s.t.} \\
		& v_c \leq \sum_{i\in P_c}y_i + \sum_{i\in N_c}(1-y_i) & c\in C_\phi \\
		& 0\leq y_i \leq 1 & \forall i L_\phi\\
		& 0\leq z_c \leq 1 & \forall c \in C_\phi
		\end{align*}
	\end{scriptsize}
	\noindent where $C_\phi$ is set of clauses of $\phi$, $L_\phi$ is the set of literals of $\phi$, $P_c$ and $N_c$ are the sets of literals appearing in clause $c$ non-negated or negated respectively, $y_i$ is the probability of setting literal $i$ to true.
	
	Consider a game $\Gamma_\phi$ encoding a generic $\phi$. If we apply the relaxation of the best-response oracle to $\Gamma_\phi$, $\mathcal{A}^R_{\textsf{BR-T}}$ and $\mathcal{A}^R_{\textsf{MAX-SAT}}$ are equivalent. To see that, first let player $2$ determine her realization plan $r^*_2$. Once $r^*_2$ has been fixed, player $1$ has, at each of her information sets, a fixed expected utility $\upsilon_a$ associated with each of her available actions $a$. Let $\{a_1, \ldots, a_k\}$ be the set of available actions at one of the information sets of player 1. There are three possible cases:
	\begin{enumerate}
		\item $\sum_{a \in \{a_1, \ldots, a_k\}}\upsilon_a < 1$.
		In this case player 1 selects, for each $a$, a probability $p(a)\geq \upsilon_a$. 
		\item $\exists a\in \{a_1, \ldots, a_k\}, \upsilon_a=1$.
		In this case, playing $a$ with probability 1 guarantees player 1 to satisfy the corresponding clause.
		\item $\sum_{a \in \{a_1, \ldots, a_k\}}\upsilon_a \geq 1$ and $\forall a\in \{a_1, \ldots, a_k\}, \upsilon_a<1$.
		In this case, $a_1$ is selected with probability $p(a_1)=\upsilon_{a_1}$, $a_2$ is selected with probability $p(a_2)=\min\{1-p(a_1), \upsilon_{a_2}\}$ and so on. The resulting strategy profile guarantees expected utility one for the corresponding clause.
	\end{enumerate} 
	Therefore, the value reachable in each clause is determined only by the choice of player 2, i.e., the final utility of the team depends only on $r_2^*$. Being the objective functions of the two formulations equivalent, the relaxed oracle enforces the same probability distribution over literals' truth assignments. That is, the optimal values of $r_2^*$ and $y^*_i$ are equivalent. Notice that, in these game instances, player 2 plays only on one level and we can sample a solution to \textsf{MAX-SAT} according to $r_2^*$ as if it was $y^*_i$. Therefore, randomized rounding of $r^*_2$ leads to the same approximation guarantee of $\mathcal{A}^R_{\textsf{MAX-SAT}}$, i.e., $(1-1/e)$~\cite{williamson2011}.\hfill$\Box$
	

\section{On forcing T-observability} Algorithm~\ref{alg:t_obs} provides a possible implementation of the procedure described in the proof of Lemma 1. Denote with $\Gamma$ the initial game tree and with $\hat{\Gamma}$ the corresponding T-observable game. The function receives as input: the current node $x$; the current sequence for the team $q$, a dictionary \textit{ids} specifying, for each player, an available information set id; a dictionary of dictionaries $d$, with the first level indexed over tuples of sequences, and the second level indexed over identifiers of information sets of $\Gamma$ (i.e., pairs player/information-set id); the id of the adversary ($adv$). By initializing $x$ to the root of the tree, the algorithm traverses the tree and assigns new information sets ids when possible, partitioning the old information structure of $\Gamma$. Once the execution is completed, $d$ can be used to create $\hat{\Gamma}$. Let us denote with $q_x$ and $h_x$ the team sequence and information set defined by decision node $x$ in $\Gamma$. For each decision-node $x$ of $\Gamma$ s.t. $\iota(x)\in T$, the corresponding decision node in $\hat{\Gamma}$ is assigned an information set with id $d[q_x][h_x]$. 

\vspace{1cm}
\setcounter{algorithm}{1}
\begin{algorithm}[!h]
	\caption{\texttt{Forcing T-Observability}}
	\begin{scriptsize}
		\begin{algorithmic}[1]
			\Function{FORCE-T-OBS}{$x$, $q$, $ids$, $d$, $adv$}\label{alg:input}
			\If{$x$ is not terminal}
				\State $pl\leftarrow x.\textnormal{player}$
				\State is\_team $\leftarrow pl\neq adv$
				\If{is\_team}
					\State $i\leftarrow (pl, x.\textnormal{id})$
					\State $q_t\leftarrow \textnormal{tuple}(q)$
					\If{$q_t$ \textbf{in} $d$}
						\If{$i$ \textbf{not in} $d[q_t]$}
							\State $d[q_t][i]\leftarrow ids[pl]$
							\State $ids[pl]\leftarrow ids[pl]+1$
						\EndIf
					\Else
						\State $d[q_t]\leftarrow \{i:ids[pl]\}$
						\State $ids[pl]\leftarrow ids[pl]+1$
					\EndIf
				\EndIf
				\For{$y$ \textbf{in} $x.$children}
					\If{is\_team}
						\State $q.$append($y$.action\_in)
					\EndIf
					\State FORCE-T-OBS($y$, $q$, $ids$, $d$, $adv$)
					\If{is\_team}
						\State $q.$pop()
					\EndIf
				\EndFor
			\EndIf
			\State \textbf{return} $d$
			\EndFunction
		\end{algorithmic}
	\end{scriptsize}
	\label{alg:t_obs}
\end{algorithm}

\newpage

\section{Additional Experimental Results}

\begin{figure*}
	\centering
	\includegraphics[width=0.8\textwidth]{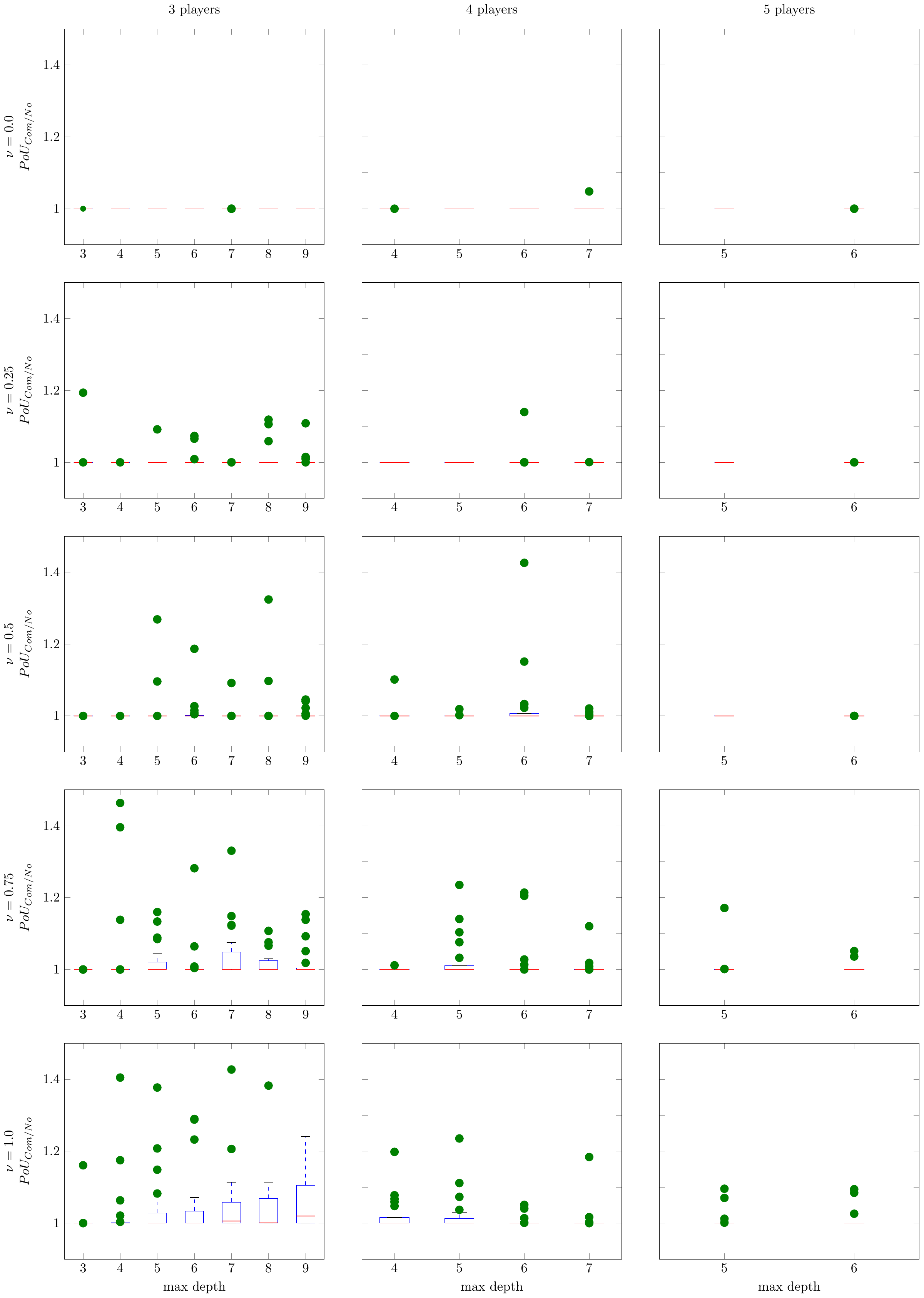}
	\caption{Box plots of the $PoU_{Com/No}$ inefficiency index.}
	\label{fig:pou_com_no}
\end{figure*}

\begin{figure*}
	\centering
	\includegraphics[width=0.8\textwidth]{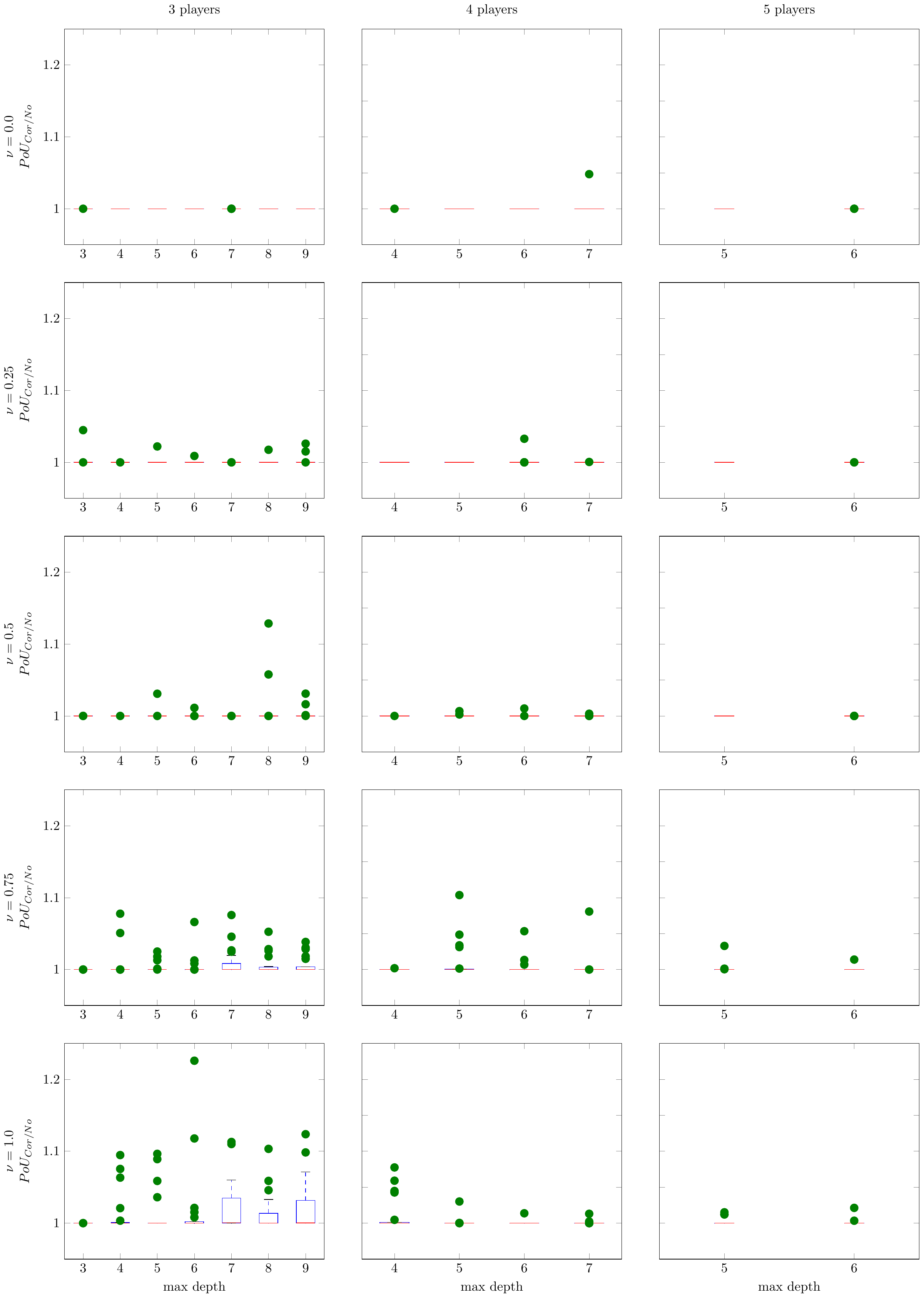}
	\caption{Box plots of the $PoU_{Cor/No}$ inefficiency index.}
	\label{fig:pou_cor_no}
\end{figure*}

\begin{figure*}
	\centering
	\includegraphics[width=0.8\textwidth]{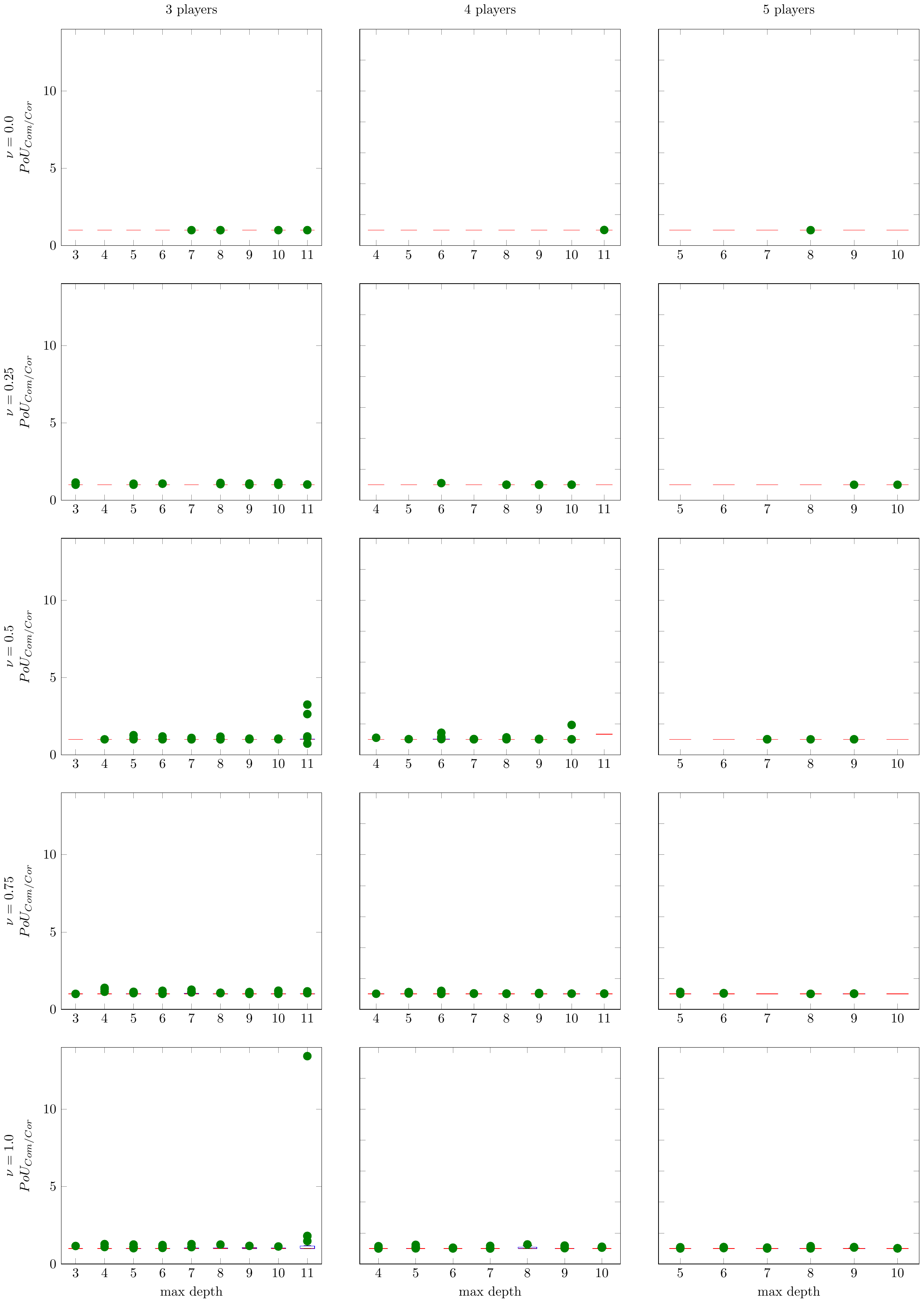}
	\caption{Boxplots of the $PoU_{Com/Cor}$ inefficiency index.}
	\label{fig:pou_com_cor}
\end{figure*}

\textbf{Empiric PoUs.} We present the box plots describing the empiric inefficiency indexes on all the experimental instances. Figure~\ref{fig:pou_com_no} describes the empiric $PoU_{Com/No}$, Figure~\ref{fig:pou_cor_no} describes the empiric $PoU_{Cor/No}$, and Figure~\ref{fig:pou_com_cor} describes the empiric $PoU_{Com/Cor}$. Notice that each plot displays data for a number of actions up to the biggest instances solvable within the time threshold by both the equilibrium-finding algorithms involved in the ratio. 

\begin{figure*}[h]
	\centering
	\includegraphics[width=0.5\textwidth]{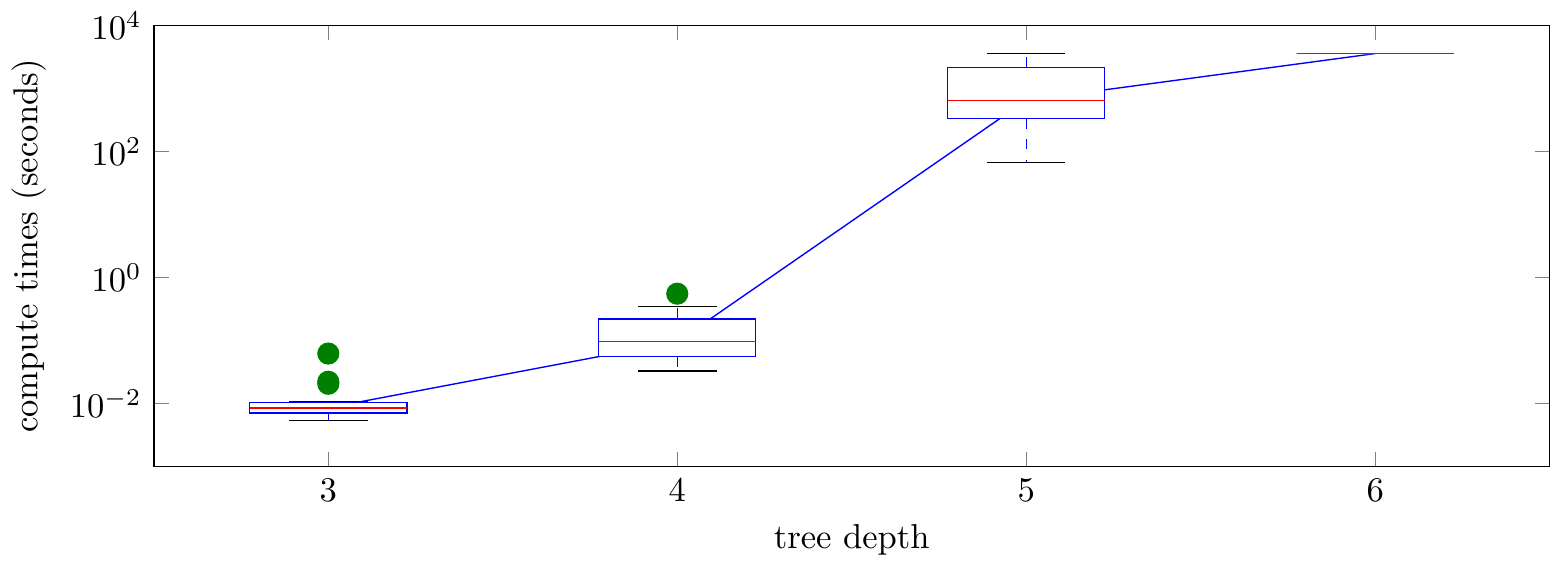}
	\caption{Average compute times and box plots of the computation of TMECor through the reduced normal form. The plot is on 3-player instances with $\nu=0.5$.}
	\label{fig:nfcor}
\end{figure*}

\textbf{TMECor in reduced normal form.} Computing a TMECor through its reduced normal form~\cite{kohlberg1986} is impractical even for relatively small game instances. Figure~\ref{fig:nfcor} shows that, for 3-player games with $\nu=0.5$, the algorithm does not reach termination within the deadline even for instances of depth 6. Moreover, the amount of memory required by the algorithm would make the computation unfeasible even with a higher time thresholds. Instances of 3-player STSA-EF-TGs with depth 6 required, at least, around $20Gb$ of memory each, with the most demanding instances requiring more than $70Gb$. Being the growth of the reduced normal form exponential in the size of the tree, this approach is not feasible for bigger game instances.

\textbf{Compute time.} Figure~\ref{fig:computetimes} shows the compute times for all the instances of our experimental setting. The behavior displayed by the equilibrium-finding algorithms is essentially the one described in the main paper for all the game configurations.

\begin{figure*}
	\centering
	\includegraphics[width=0.8\textwidth]{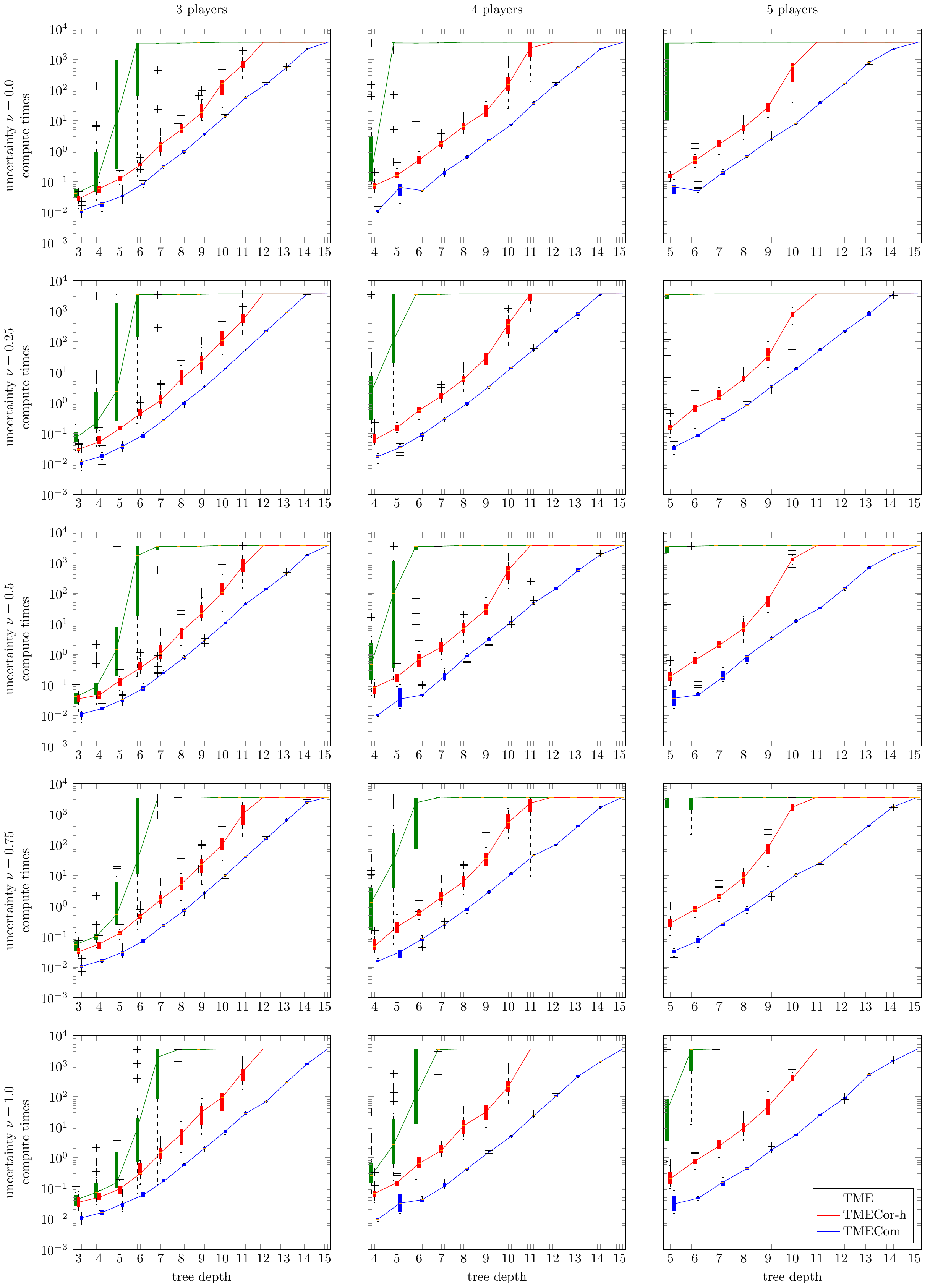}
	\caption{Average compute times of the algorithms and their box plots with every game configuration.}
	\label{fig:computetimes}
\end{figure*}

\end{document}